\documentclass[sigconf]{acmart}

\usepackage{colortbl}
\usepackage{xcolor} 
\usepackage[flushleft]{threeparttable}
\usepackage{colortbl}
\usepackage{makecell} 
\usepackage{tabularx}
\usepackage{multirow}
\usepackage{bm}
\usepackage{algorithm}
\usepackage{algpseudocode}
\usepackage{makecell}
\usepackage{balance}
\newcommand\STAB[1]{\begin{tabular}{@{}c@{}}#1\end{tabular}}

\definecolor{mygreen}{RGB}{10,200,160}
\definecolor{myblue}{RGB}{31,78,121}
\definecolor{myyellow}{RGB}{191,144,0}
\definecolor{myorange}{RGB}{197,90,17}

\AtBeginDocument{%
  }

\copyrightyear{2025}
\acmYear{2025}
\setcopyright{acmlicensed}
\acmConference[MM '25]{Proceedings of the 33rd ACM International Conference on Multimedia}{October 27--31, 2025}{Dublin, Ireland}
\acmBooktitle{Proceedings of the 33rd ACM International Conference on Multimedia (MM '25), October 27--31, 2025, Dublin, Ireland}
\acmDOI{10.1145/3746027.3755359}
\acmISBN{979-8-4007-2035-2/2025/10}

\begin{document}

\title{Regularizing Subspace Redundancy of Low-Rank Adaptation}

\author{Yue Zhu}
\authornote{Equal Contribution}
\affiliation{%
  \institution{Dalian University of Technology}
  \city{Dalian}
  \country{China}
}
\email{zhuyuedlut@gmail.com}

\author{Haiwen Diao}
\authornotemark[1]
\affiliation{%
  \institution{Dalian University of Technology}
  \city{Dalian}
  \country{China}
}
\email{diaohw@mail.dlut.edu.cn}

\author{Shang Gao}
\authornotemark[1]
\affiliation{%
  \institution{Dalian University of Technology}
    \city{Dalian}
  \country{China}
}
\email{gs940601k@mail.dlut.edu.cn}

\author{Jiazuo Yu}
\affiliation{%
  \institution{Dalian University of Technology}
    \city{Dalian}
  \country{China}
}
\email{yujiazuo@mail.dlut.edu.cn}

\author{Jiawen Zhu}
\affiliation{%
  \institution{Dalian University of Technology}
    \city{Dalian}
  \country{China}
}
\email{jiawen@mail.dlut.edu.cn}

\author{Yunzhi Zhuge}
\affiliation{%
  \institution{Dalian University of Technology}
    \city{Dalian}
  \country{China}
}
\email{zgyz@dlut.edu.cn}

\author{Shuai Hao}
\affiliation{%
  \institution{Dalian University of Technology}
    \city{Dalian}
  \country{China}
}
\email{shuaihao@mail.dlut.edu.cn}

\author{Xu Jia}
\affiliation{%
  \institution{Dalian University of Technology}
    \city{Dalian}
  \country{China}
}
\email{xjia@dlut.edu.cn}

\author{Lu Zhang}
\affiliation{%
  \institution{Dalian University of Technology}
    \city{Dalian}
  \country{China}
}
\email{zhanglulu@dlut.edu.cn}

\author{Ying Zhang}
\affiliation{%
  \institution{WeChat Vision, Tencent Inc.}
    \city{Dalian}
  \country{China}
}
\email{yinggzhang@tencent.com}

\author{Huchuan Lu}
\authornote{Corresponding Authors}
\affiliation{%
  \institution{Dalian University of Technology}
    \city{Dalian}
  \country{China}
}
\email{lhchuan@mail.dlut.edu.cn}

\renewcommand{\shortauthors}{Yue Zhu, Haiwen Diao, Shang Gao et al.}

\begin{abstract}
Low-Rank Adaptation (LoRA) and its variants have delivered strong capability in Parameter-Efficient Transfer Learning (PETL) by minimizing trainable parameters and benefiting from reparameterization. 
However, their projection matrices remain unrestricted during training, causing high representation redundancy and diminishing the effectiveness of feature adaptation in the resulting subspaces.
While existing methods mitigate this by manually adjusting the rank or implicitly applying channel-wise masks, they lack flexibility and generalize poorly across various datasets and architectures.
Hence, we propose \textbf{ReSoRA}, a method that explicitly models redundancy between mapping subspaces and adaptively \textbf{Re}gularizes \textbf{S}ubspace redundancy of L\textbf{o}w-\textbf{R}ank \textbf{A}daptation. 
Specifically, it theoretically decomposes the low-rank submatrices into multiple equivalent subspaces and systematically applies de-redundancy constraints to the feature distributions across different projections.
Extensive experiments validate that our proposed method consistently facilitates existing state-of-the-art PETL methods across various backbones and datasets in vision-language retrieval and standard visual classification benchmarks. 
Besides, as a training supervision, \textbf{ReSoRA} can be seamlessly integrated into existing approaches in a plug-and-play manner, with no additional inference costs.
Code is publicly available at: \href{https://github.com/Lucenova/ReSoRA}{\textcolor{blue}{https://github.com/Lucenova/ReSoRA}}.
\end{abstract}

\begin{CCSXML}
<ccs2012>
   <concept>
    <concept_id>10010147.10010257.10010321.10010337</concept_id>
       <concept_desc>Computing methodologies~Transfer learning</concept_desc>
       <concept_significance>500</concept_significance>
       </concept>
 </ccs2012>
\end{CCSXML}

\ccsdesc[500]{
Computing methodologies~Transfer learning}

\keywords{Parameter-efficient transfer learning; low-rank adaption; subspace regularization; plug-and-play}

\maketitle

\begin{figure}[htp]
    \centering
    \includegraphics[width=\linewidth]{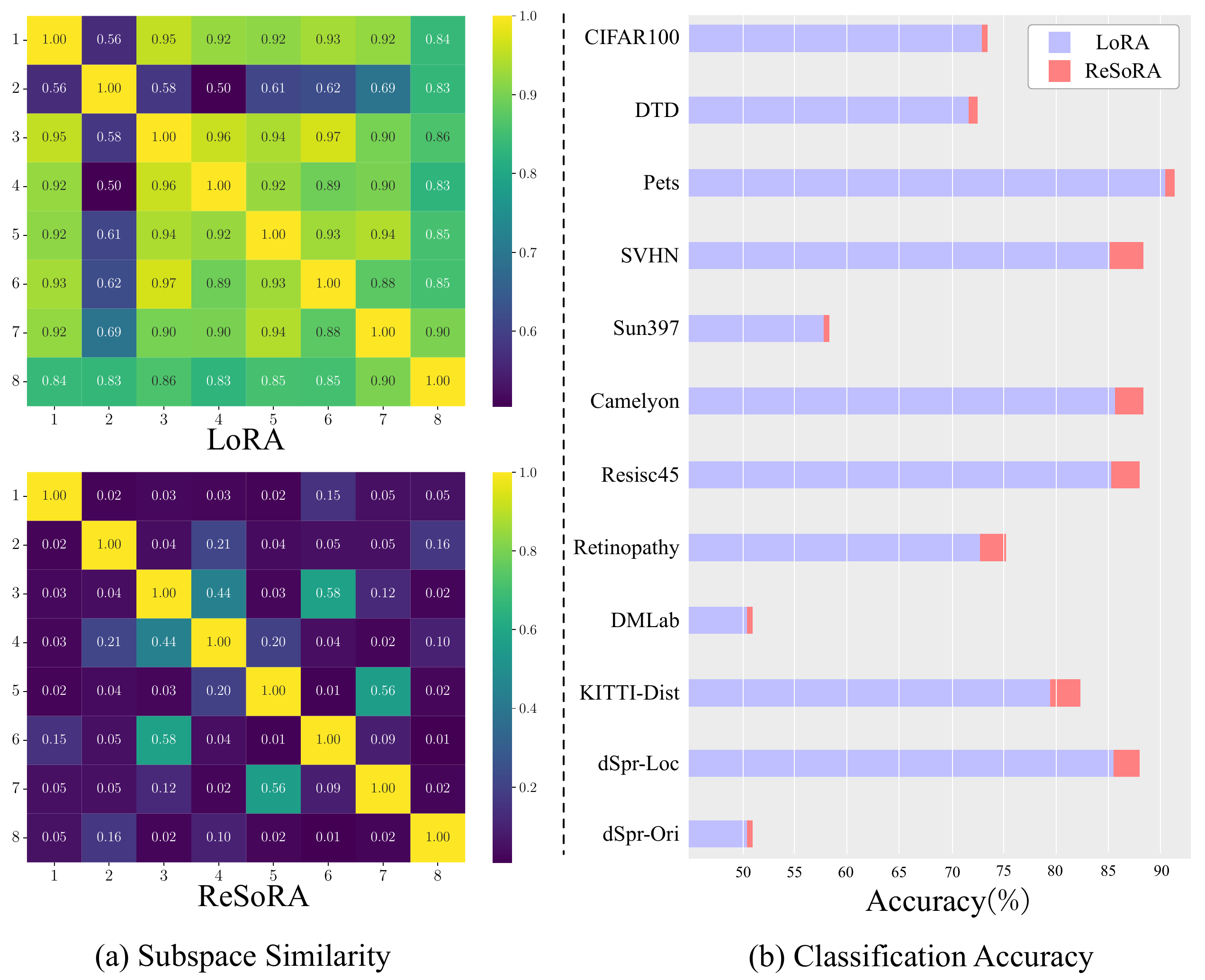}
    \caption{Comparison between our ReSoRA and LoRA~\cite{TL:LoRA}: (a) subspace redundancy regularization over decoupled query matrix in the last ViT layer on SVHN; (b) prediction accuracy improvements across various classification benchmarks.}
    \label{fig:cka_coefficient}
\end{figure}

\section{Introduction}
Recently, large fundamental models~\cite{TransF:ViT,TransF:GPT,VLP:CLIP,VLM:EVE,VLM:EVEv2} have exhibited remarkable generalization capabilities across natural language processing (NLP), computer vision (CV), and vision-language (VL) tasks.
Nevertheless, fine-tuning all network parameters is computationally and storage-intensive, and with limited data scales, it may impede adaptation and result in suboptimal performance.

To address this, Parameter-Efficient Transfer Learning (PETL) techniques~\cite{TL:VPT,TL:LoRA,TL:AdaptFormer} have been increasingly gaining attention across a range of downstream applications~\cite{TL:PathWeave,TL:SHERL}. 
They typically freeze most model parameters, either inserting small trainable blocks~\cite{TL:Adapter-BERT,TL:ViT-Adapter,TL:VL-ADAPTER} into the pre-trained model or selectively adjusting a limited subset of parameters~\cite{TL:BitFit,TL:Layernorm-Tuning, TL:DiffPruning}. They significantly reduce trainable parameters while preserving or improving model capability. Among them, Low-Rank Adaptation (LoRA)~\cite{TL:LoRA,TL:AdaLoRA,TL:DTL,TL:FacT} stands out for its simplicity and efficiency.
It assumes that weight updates during the fine-tuning process have a low intrinsic rank and adapts by optimizing the corresponding decomposition matrices.
During inference, the low-rank matrices can be integrated seamlessly into the pre-trained backbone weights through reparameterization, incurring no additional prediction latency.

Although effective, LoRA shows significant redundancy in its low-rank adaptation matrices. To quantify it, we decompose the mapping function into multiple subspace components, where high inter-subspace redundancy means more overlap and less diversity. Figure~\ref{fig:cka_coefficient} shows strong correlations between LoRA’s subspaces, indicating poor disentanglement. Combined with limited gains, this highlights the need for improved training strategies that suppress similarity and encourage orthogonality across projection spaces.

Recent studies~\cite{TL:AutoLoRA,TL:AdaLoRA,TL:SoRA} aim to build more compact and efficient mapping spaces. AdaLoRA~\cite{TL:AdaLoRA} applies SVD to reformulate adaptation matrices and prunes singular values by importance, while SoRA~\cite{TL:SoRA} uses a gated mechanism with proximal gradient descent and $\ell_1$ regularization to remove inactive components. Although these approaches reduce redundancy and improve efficiency, they remain constrained by fixed ranks and show limited flexibility and generalization across datasets and architectures.

In this paper, we present ReSoRA that decomposes the adaptive output space into multiple subspaces and introduces a tailored subspace regularization term. We systematically identify the most effective regularization strategies to enforce mapping constraints and reduce redundancy across subspaces, improving representational capacity and enabling better domain adaptation.

Our contributions can be summarized as follows:
\begin{itemize}
\item We explicitly decompose the low-rank mapping space into multiple subspaces and effectively regularize redundancy to promote orthogonality among them, offering a new perspective on facilitating low-rank adaptation.
\item We introduce ReSoRA, a simple yet effective method that serves as a plug-and-play anti-redundancy regularizer during training. 
It consistently enhances various state-of-the-art PETL works, e.g., LoRA~\cite{TL:LoRA}, FacT~\cite{TL:FacT}, MoSLoRA~\cite{TL:MSLoRA}, and DTL~\cite{TL:DTL}, without any inference overhead.

\item We perform extensive experiments on various backbones and datasets in visual language retrieval and standard visual classification benchmarks, demonstrating the effectiveness and robustness of the proposed ReSoRA. 
\end{itemize}

\begin{figure*}[htp]
    \centering
    \includegraphics[width=0.85\linewidth]{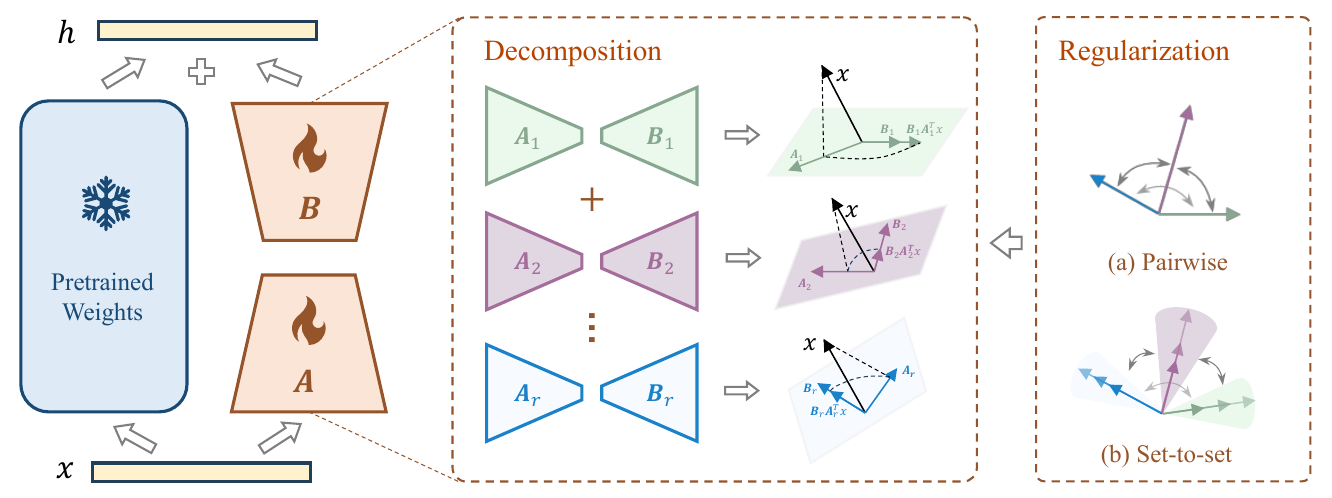}
    \caption{Overview of our proposed ReSoRA. The original output space is decomposed into a combination of multiple subspaces. To reduce subspace redundancy, we adopt a carefully designed regularization that includes pairwise and set-to-set terms.}
    \label{fig:pipeline}
\end{figure*}

\section{Related Work}
\label{sec:Related-work}
\subsection{Parameter-Efficient Transfer Learning}
Parameter-Efficient Transfer Learning (PETL)~\cite{TL:Prefix-Tuning,TL:KARST} has been developed to selectively fine-tune a small subset of parameters, achieving high efficiency.
Among them, partial tuning approaches~\cite{TL:BitFit,TL:Layernorm-Tuning} adapt pre-trained models by training only a subset of parameters. BitFit~\cite{TL:BitFit} updates only bias terms, while LayerNorm Tuning~\cite{TL:Layernorm-Tuning} adjusts parameters within Layer Normalization layers. 
Besides, prompt-tuning methods~\cite{TL:VPT,TL:Efficient-Prompt} integrate trainable prompt tokens into the input sequence, enabling task-specific learning without modifying pretrained weights. 
Furthermore, adapter-based methods~\cite{TL:CLIP-Adapter,ITM:GSSF} insert lightweight modules within transformers for adaptation. 
Notably, Adapter Re-composing~\cite{TL:AdaReComposing}, RLRR~\cite{TL:RLRR}, and Householder Adaptation~\cite{TL:Householder} achieve strong performance by improving parameter efficiency via multiple adapter compositions, residual low-rank tuning, and Householder transformations.

A notable PETL method is LoRA~\cite{TL:LoRA}, which models the weight update matrix $\Delta{W}$ in a linear layer as a low-rank approximation. Building on this, subsequent studies have proposed advanced techniques to further improve adaptation, including factorized bilinear matrices~\cite{TL:FacT}, compactor decomposition~\cite{TL:Compacter}, and tensor factorization~\cite{TL:superLoRA}. Other works~\cite{TL:DTL} decouple weight updates from the backbone via a lightweight side network. Our ReSoRA naturally complements this paradigm, integrating seamlessly with various LoRA-like methods while adding no inference overhead.

\subsection{Regularization in Low-Rank Adaptation}
Low-rank adaptation reduces the number of trainable parameters by restricting weight updates to rank-$r$ subspaces. While commonly used ranks ($r = 8$, $16$, $32$) are typically chosen based on empirical heuristics, the optimal rank can vary with task complexity and requirements. Selecting an appropriate $r$ is crucial—an excessively high rank may introduce redundancy within low-rank matrices, leading to potential inefficiency and performance degradation.

To address it, one group of methods focuses on selecting an appropriate rank to balance expressiveness and efficiency. For instance, IncreLoRA~\cite{TL:IncreLoRA} incrementally increases the rank based on parameter importance scores. AdaLoRA~\cite{TL:AdaLoRA} adopts Singular Value Decomposition (SVD) to parameterize rank adaptation, while AutoLoRA~\cite{TL:AutoLoRA} leverages meta-learning to adjust ranks automatically during training.
Besides, another group aims to enhance the independence and diversity of low-rank matrices. Among them, MLAE~\cite{TL:MLAE} introduces a cellular decomposition strategy that factorizes a low-rank matrix into independent rank-1 components. SiRA~\cite{TL:SiRA} employs a Sparse Mixture of Experts (SMoE) to boost LoRA's capacity, and MoSLoRA~\cite{TL:MSLoRA} uses a learnable mixer to more flexibly combine diverse subspaces.
Orthogonally, ReSoRA decomposes low-rank submatrices into equivalent subspaces and enforces de-redundancy constraints across their feature distributions, which can seamlessly cooperate with various LoRA-like methods during fine-tuning.

\section{Methodology}
\label{sec:Methodology}

In this paper, we present ReSoRA, a general and adaptive framework designed to enhance low-rank adaptation. At its core is subspace redundancy regularization, a family of techniques that explicitly mitigates redundancy and promotes representation diversity across decomposed subspaces. Our ReSoRA is flexible and extensible, supporting both pairwise or set-to-set regularization items.

\subsection{Subspace Projection Decomposition}

To demonstrate the effectiveness of our ReSoRA, we focus on enhancing the widely-used LoRA~\cite{TL:LoRA,TL:MSLoRA} method by incorporating ReSoRA as regularization. 
Concretely, LoRA assumes that weight updates during fine-tuning occur in a low-dimensional subspace and approximates these updates as the product of two low-rank matrices.
Formally, given a pre-trained weight matrix $W_0 \in \mathbb{R}^{d_{in} \times d_{out}}$, where $d_{in}$ and $d_{out}$ denote the input and output dimensions respectively, the fine-tuned weight $W$ can be expressed as:
\begin{equation}
W = W_0 + \Delta{W} = W_0 + BA,
\label{formation:lora}
\end{equation}
where $BA$ is a low-rank decomposition of the adaptive weight increment $\Delta{W}$. Hence $B\in \mathbb{R}^{d_{in} \times r}$ and $A\in{\mathbb{R}^{r \times d_{out}}}$, with $r \ll min\{d_{in}, d_{out}\}$.
During the training phase, the pre-trained weight $W_0$ remains unchanged, while the parameters in $A$ and $B$ are trainable. 
Subsequently, $\Delta{W}$ can be merged into $W_0$ without introducing any latency during inference. 
To investigate the redundancy in low-rank adaptation, we first perform decomposition on the parameter $\Delta{W}$.
Specifically, both $A$ and $B$ can be decomposed into rank-1 subspaces $r$: $A=[A_1, A_2, \cdots, A_r]^{T}$, $B = [B_1, B_2, \cdots, B_r]$, where $A_i\in\mathbb{R}^{d_{in}\times 1}$ and $B_i\in\mathbb{R}^{d_{out} \times 1}$ for $1 \leq i \leq r$.
Thus, the adaptive weight in Eq.~\eqref{formation:lora} can be explicitly decomposed as: 
\begin{align}
W &= W_0 + W_1 + W_2 + \cdots + W_r  \nonumber \\ 
  &= W_0 + B_1A_1^T + B_2A_2^T + \cdots + B_rA_r ^T,
\label{formation:lora-decompose}
\end{align}
where $W_i=B_iA_i^T$ is a rank-1 matrix derived from the product of the corresponding submatrices $B_i$ and $A_i$.

\subsection{Subspace Redundancy Formulation}

Given the input $x \in \mathbb{R}^{d_{in}}$, the final output $h \in \mathbb{R}^{d_{out}}$ of the adapted module can be represented as:
\begin{equation}
   h = h_0 + \Delta{h} = W_0x + BAx.
\label{formation:h-decompose}
\end{equation}

\begin{table*}[t]\small
    \caption{Performance improvements on Flickr30K using \emph{VSE$\infty$} with single Transformer {\scriptsize(BERT + Region Features)}, and MSR-VTT using \emph{CLIP4Clip} with dual Transformer encoders {\scriptsize(ViT + Text Transformer)}. 
    We report Recall@1 (R@1) on sentence retrieval (``I-T'', ``V-T''), image retrieval (``T-I''), video retrieval (``T-V''), and ``Rsum'' of R@1,5,10 on bi-directional retrievals.}
    \centering
    \begin{tabularx}{\textwidth}{
        >{\raggedright\arraybackslash}X 
        *{5}{>{\centering\arraybackslash}X}
        | 
        *{5}{>{\centering\arraybackslash}X} 
    }
    \toprule
    \multirowcell{2}{Method} 
    & Params. 
    & Memory 
    &\multicolumn{3}{c|}{BERT + Region Features} 
    & Params. 
    & Memory 
    &\multicolumn{3}{c}{ViT + Text Transformer} \\
    \cmidrule{4-6} \cmidrule{9-11}
    &(M) &(G) 
    &I-T &T-I &Rsum 
    &(M) &(G) 
    &T-V &V-T &Rsum\\
    \midrule
    Full
    &109.5 &9.9
    &79.7 & 62.1 & 513.5
    &151.3 & 12.2 * 4
    &42.8 &42.1 &389.2 
    \\
    Partially
    &0.8 &1.0
    &74.8 &53.7 &485.5
    &0.7 &1.9 * 4
    &36.4 &37.0 &353.9
    \\
    \midrule
    Adapter~\cite{TL:Adapter-BERT}
    &2.6 &8.8 
    &79.1 &60.5 &511.3
    &5.2 &10.3 * 4
    &38.3 &39.6 &364.3
    \\
    BitFit~\cite{TL:BitFit} 
    &0.9 &8.6 
    &77.3 &57.8 &503.9
    &0.1 &10.5 * 4
    &38.1 &40.6 &370.8
    \\    
    Prompt~\cite{TL:Prefix-Tuning}
    &10.7 &9.4 
    &78.7 &59.0 &508.5
    &0.2 &10.7 * 4
    &36.8 &37.5 &358.8
    \\
    SSF~\cite{TL:SSF}
    &0.2 &8.4
    &80.0 &60.4 &512.8
    &0.5 &9.8 * 4
    &40.2 &41.8 &376.6
    \\
    FacT~\cite{TL:FacT}
    &0.6 &8.7
    &79.2 &59.3 &508.8
    &0.8 &10.2 * 4
    &38.7 &39.8 &367.2 
    \\
    LST~\cite{TL:LST}
    &7.5 &4.6 
    &77.9 &57.3 &501.9
    &11.2 &8.0 * 4
    &37.0 &37.8 &356.7
    \\    
    UniPT~\cite{TL:UniPT}
    &5.9 &3.1 
    &80.2 &59.8 &510.5
    &9.6 &3.4 * 4
    &38.9 &39.3 &361.3
    \\
    AdaLoRA~\cite{TL:AdaLoRA}
    &1.0 &8.8
    &79.8 &60.1 &510.3
    &1.2 &10.5 * 4
    &39.2 &39.6 &368.5 
    \\
    \rowcolor{gray!20}
    LoRA~\cite{TL:LoRA}
    &1.1 &8.8 
    &78.8 &59.6 &508.2
    &1.3 &10.2 * 4
    &38.8 &39.9 &366.8
    \\
    \; + ReSoRA
    & 1.1  & 9.8 
    & \textcolor{mygreen}{1.0~$\uparrow$} 
    & \textcolor{mygreen}{0.8~$\uparrow$}
    & \textcolor{mygreen}{2.6~$\uparrow$}
    & 1.3 & 11.8 * 4
    & \textcolor{mygreen}{0.6~$\uparrow$}
    & \textcolor{mygreen}{0.6~$\uparrow$}
    & \textcolor{mygreen}{3.3~$\uparrow$}
    \\
    \bottomrule
    \end{tabularx}
    \label{tab:VL_PETL}
\end{table*}

We focus on the incremental term $\Delta{h}$, as it contains all trainable parameters from LoRA. 
Following Eq.~\eqref{formation:lora-decompose}, the output increment can be decomposed into a sum of rank-1 components as follows:
\begin{align}
  \Delta{h} &= BAx \nonumber \\
            &= B_1A_1^Tx + B_2A_2^Tx + \cdots + B_rA_r^Tx \nonumber \\
            &= \Delta{h_1} + \Delta{h_2} + \cdots + \Delta{h_r}.
\label{formation:delta_h}
\end{align}
Here, each sub-feature $\Delta h_i = B_i A_i^T x$ represents the contribution of the $i$-th rank-1 subspace, and $A_i$, $B_i$ are the $i$-th column and row components from the low-rank matrices $A$ and $B$, respectively.

As shown in Figure~\ref{fig:pipeline}, the full update $\Delta h$ is essentially a linear combination of the vectors $B_i$, modulated by the inner products between $x$ and each $A_i$. 
This formulation allows us to interpret $\Delta h$ as a composition of multiple subspace projections. 
To improve the expressiveness of the model, it is desirable for these subspaces to exhibit distinct characteristics. 
However, in practice, many of the rank-1 components are highly correlated, leading to redundancy and reduced representational diversity. 
We refer to this phenomenon as \textbf{feature-level redundancy} across multiple subspaces.

\subsection{Subspace Regularization Formulation}

To mitigate subspace redundancy, we propose a general subspace regularization term $R(B, A, X)$, where $X \in \mathbb{R}^{d_{in} \times N}$ represents a mini-batch of $N$ input samples. ReSoRA independently applies regularization to each LoRA component, explicitly penalizing redundancy among subspaces, thereby effectively enhancing the model's adaptation capacity. Specifically, we introduce two categories of regularization methods: pairwise and set-to-set. Pairwise regularization evaluates redundancy between feature vectors, while set-to-set regularization assesses redundancy at the feature-set level, capturing structural overlaps beyond point-wise comparisons.

\noindent\textbf{Pairwise Regularization.} We first adopt commonly-used \textit{Euclidean distance} to quantify redundancy among incremental features projected onto subspaces, denoted by $\Delta{h_i}$:
\begin{align}
R_e(B, A, X) &= \frac{1}{N}\sum_{n=1}^{N}\sum_{i=1}^{r}\sum_{j=i+1}^{r} e^{-\beta \Vert \Delta{h^n_i}-\Delta{h^n_j} \Vert^2_2},
\end{align}
where $\Delta{h^n_i}$ denotes the incremental output of the $n$-th sample projected onto the $i$-th subspace, and $\beta$ is a scaling factor controlling feature-space sensitivity, enhancing robustness to noise.

Furthermore, we employ \textit{Cosine distance} to capture directional differences independent of vector magnitude as follows:
\begin{align}
R_c(B, A, X) = \frac{1}{N}\sum_{n=1}^{N}\sum_{i=1}^{r}\sum_{j=i+1}^{r} \frac{(\Delta{h^n_i})^T\Delta{h^n_j}}{\Vert\Delta{h^n_i}\Vert_2 \Vert\Delta{h^n_j}\Vert_2}.
\end{align}

\noindent\textbf{Set-to-set Regularization.} 
Unlike pairwise items, set-to-set items capture global alignment and correlation across entire feature sets. We explore two well-designed regularization strategies to quantify structural similarities between different subspaces.

Specifically, \textit{Linear measurement} (default setting) is defined as:
\begin{equation}
R_{l}(B, A, X) = \sum_{i=1}^{r}\sum_{j=i+1}^{r} \frac{\Vert \Delta{H_i}\Delta{H_j}^T \Vert_F^2}{\Vert \Delta{H_i} \Delta{H_i}^T \Vert_F \Vert \Delta{H_j} \Delta{H_j^T} \Vert_F},
\end{equation}
where $\Delta{H_i} \in \mathbb{R}^{d_{in} \times N}$ denotes batch features extracted from the $i$-th subspace, and $|\cdot|_F$ is the Frobenius norm.

Besides, we propose a \textit{Nonlinear measurement} based on a kernel function to capture more complex interactions:
\begin{equation}
R_{n}(B, A, X) = \sum_{i=1}^{r}\sum_{j=i+1}^{r} \frac{\text{tr}(K_i M K_j M)}{\sqrt{\text{tr}((K_i M)^2)\text{tr}((K_j M)^2)}},
\end{equation}
where $M = I_N-\frac{1}{N}\bm{1}\bm{1}^T$ is a centering matrix, and $K_i(p,q) = \exp(-\Vert \Delta{h^p_i} - \Delta{h^q_i}\Vert_2^2/(2\sigma^2))$ with variance parameter $\sigma$ determined as a fraction of the median sample distances.
These designs can explicitly suppress redundancy across distinct subspaces, thus substantially enhancing feature diversity and improving generalization performance in LoRA-based fine-tuning methods.

\noindent\textbf{Theoretical Justification.} 
Here we revisit the Cosine distance based regularization from Eq.~(6), simplified here as:
\begin{equation}
R_c(B, A, X) = \sum_{i,j} \frac{(\Delta h_i)^\top \Delta h_j}{\Vert\Delta h_i\Vert_2 \Vert\Delta h_j\Vert_2}.
\end{equation}
The gradient of $R_c$ with respect to $\Delta h_i$ is derived as follows:
\begin{equation}
\frac{\partial R_c}{\partial \Delta h_i} = \sum_{j}\frac{1}{\Vert\Delta h_i\Vert_2\Vert\Delta h_j\Vert_2} \left(\Delta h_j - \cos(\theta_{ij}) \frac{\Delta h_i}{\Vert\Delta h_i\Vert_2}\right)
\end{equation}
where $\cos(\theta_{ij})$ denotes cosine similarity between vectors $\Delta h_i$ and $\Delta h_j$. 
This gradient explicitly penalizes directional similarity between subspace projections. When $\Delta h_i$ and $\Delta h_j$ are highly aligned ($\cos(\theta_{ij}) \approx 1$), it pushes them apart, encouraging orthogonalization and reducing redundancy. This fosters feature diversity and robustness, improving generalization across tasks and domains.

\begin{table*}[thp]
\caption{Performance improvements on VTAB-1K using ViT-B/16 pre-trained on supervised ImageNet dataset.}
\centering
\setlength{\tabcolsep}{0.3pt}
\small
\begin{tabular}{p{2.2cm}<{}p{0.65cm}<{\centering}|p{0.65cm}<{\centering}p{0.65cm}<{\centering}p{0.65cm}<{\centering}p{0.65cm}<{\centering}p{0.65cm}<{\centering}p{0.65cm}<{\centering}p{0.65cm}<{\centering}|p{0.65cm}<{\centering}p{0.65cm}<{\centering}p{0.65cm}<{\centering}p{0.65cm}<{\centering}|p{0.65cm}<{\centering}p{0.65cm}<{\centering}p{0.65cm}<{\centering}p{0.65cm}<{\centering}p{0.65cm}<{\centering}p{0.65cm}<{\centering}p{0.65cm}<{\centering}p{0.65cm}<{\centering}|p{0.65cm}<{\centering}p{0.65cm}<{\centering}}
	\toprule[1.5pt]
	\multicolumn{2}{c|}{}&\multicolumn{7}{c|}{\textbf{Natural}}&\multicolumn{4}{c|}{\textbf{Specialized}}&\multicolumn{8}{c|}{\textbf{Structured}}&\\
	&\multicolumn{1}{c|}{\STAB{\rotatebox[origin=c]{90}{param (M)}}}
	&\multicolumn{1}{c}{\STAB{\rotatebox[origin=c]{90}{Cifar100}}}
	&\multicolumn{1}{c}{\STAB{\rotatebox[origin=c]{90}{Caltech101}}}
	&\multicolumn{1}{c}{\STAB{\rotatebox[origin=c]{90}{DTD}}}
	&\multicolumn{1}{c}{\STAB{\rotatebox[origin=c]{90}{Flower102}}}
	&\multicolumn{1}{c}{\STAB{\rotatebox[origin=c]{90}{Pets}}}
	&\multicolumn{1}{c}{\STAB{\rotatebox[origin=c]{90}{SVHN}}}
	&\multicolumn{1}{c|}{\STAB{\rotatebox[origin=c]{90}{Sun397}}}
	&\multicolumn{1}{c}{\STAB{\rotatebox[origin=c]{90}{Camelyon}}}
	&\multicolumn{1}{c}{\STAB{\rotatebox[origin=c]{90}{EuroSAT}}}
	&\multicolumn{1}{c}{\STAB{\rotatebox[origin=c]{90}{Resisc45}}}
	&\multicolumn{1}{c|}{\STAB{\rotatebox[origin=c]{90}{Retinopathy}}}
	&\multicolumn{1}{c}{\STAB{\rotatebox[origin=c]{90}{Clevr-Count}}}
	&\multicolumn{1}{c}{\STAB{\rotatebox[origin=c]{90}{Clevr-Dist}}}
	&\multicolumn{1}{c}{\STAB{\rotatebox[origin=c]{90}{DMLab}}}
	&\multicolumn{1}{c}{\STAB{\rotatebox[origin=c]{90}{KITTI-Dist}}}
	&\multicolumn{1}{c}{\STAB{\rotatebox[origin=c]{90}{dSpr-Loc}}}
	&\multicolumn{1}{c}{\STAB{\rotatebox[origin=c]{90}{dSpr-Ori}}}
	&\multicolumn{1}{c}{\STAB{\rotatebox[origin=c]{90}{sNORB-Azim}}}
	&\multicolumn{1}{c|}{\STAB{\rotatebox[origin=c]{90}{sNORB-Ele}}}
        &\multicolumn{1}{c}{\STAB{\rotatebox[origin=c]{90}{Average}}}
	&\multicolumn{1}{c}{\STAB{\rotatebox[origin=c]{90}{All Set Average}}}\\
	\specialrule{0em}{1pt}{1pt}
	\hline
	\specialrule{0em}{1pt}{1pt}
	\multicolumn{22}{l}{Traditional Fine-Tuning}\\
	\hline
	\specialrule{0em}{1pt}{1pt}
	Full&85.8&68.9&87.7&64.3&97.2&86.9&87.4&38.8&79.7&95.7&84.2&73.9&56.3&58.6&41.7&65.5&57.5&46.7&25.7&29.1&68.9 &65.6 \\
	Linear&0&64.4&85.0&63.2&97.0&86.3&36.6&51.0&78.5&87.5&68.5&74.0&34.3&30.6&33.2&55.4&12.5&20.0&9.6&19.2&57.6 & 53.0\\
	\hline
	\specialrule{0em}{1pt}{1pt}
	\multicolumn{22}{l}{PETL methods}\\
	\hline
	\specialrule{0em}{1pt}{1pt}
    BitFit~\cite{TL:BitFit}&0.10&72.8&87.0&59.2&97.5&85.3&59.9&51.4&78.7&91.6&72.9&69.8&61.5&55.6&32.4&55.9&66.6&40.0&15.7&25.1&65.2 &62.0\\
    VPT~\cite{TL:VPT}&0.56&78.8&90.8&65.8&98.0&88.3&78.1&49.6&81.8&96.1&83.4&68.4&68.5&60.0&46.5&72.8&73.6&47.9&32.9&37.8&72.0 & 69.4 \\
	LST~\cite{TL:LST}&2.38&59.5&91.5&69.0&99.2&89.9&79.5&54.6&86.9&95.9&85.3&74.1&81.8&61.8&52.2&81.0&71.7&49.5&33.7&45.2&74.3 &71.7\\

    AdaLoR~\cite{TL:AdaLoRA}&0.44&52.1&89.1&68.9&96.8&88.2&79.5&53.6&86.5&96.1&84.4&75.6&83.0&64.1&55.9&81.6&86.4&52.7&34.4&43.5&74.6 &72.2 \\
    AdaptFormer~\cite{TL:AdaptFormer}&0.16&70.8&91.2&70.5&99.1&90.9&86.6&54.8&83.0&95.8&84.4&76.3&81.9&64.3&49.3&80.3&76.3&45.7&31.7&41.1&74.7 &72.3 \\
    NOAH~\cite{TL:NOAH}&0.43&69.6&92.7&70.2&99.1&90.4&86.1&53.7&84.4&95.4&83.9&75.8&82.8&68.9&49.9&81.7&81.8&48.3&32.8&44.2&75.5 & 73.2\\
    SSF~\cite{TL:SSF}&0.21&69.0&92.6&75.1&99.4&91.8&90.2&52.9&87.4&95.9&87.4&75.5&75.9&62.3&53.3&80.6&77.3&54.9&29.5&37.9&75.7 &73.2 \\	

    \rowcolor{gray!20}
    FacT-TK$_{r\leq{32}}$~\cite{TL:FacT}&0.07&70.8&92.0&68.8&98.9&89.9&88.6&53.9&85.1&95.8&83.9&75.4&82.7&67.4&50.1&80.9&80.4&45.3&32.8&42.3&75.2 & 72.9 \\

    \quad + ReSoRA&0.07&\textcolor{mygreen}{2.1~$\uparrow$}&\textcolor{mygreen}{0.2~$\uparrow$}&\textcolor{mygreen}{1.4~$\uparrow$}&\textcolor{mygreen}{0.1~$\uparrow$}&\textcolor{mygreen}{0.2~$\uparrow$}&\textcolor{mygreen}{0.3~$\uparrow$}&\textcolor{mygreen}{0.4~$\uparrow$}&\textcolor{mygreen}{0.3~$\uparrow$}&\textcolor{mygreen}{0.3~$\uparrow$}&\textcolor{mygreen}{0.6~$\uparrow$}&0.0&\textcolor{mygreen}{0.2~$\uparrow$}&\textcolor{mygreen}{1.0~$\uparrow$}&\textcolor{mygreen}{1.0~$\uparrow$}&\textcolor{mygreen}{0.7~$\uparrow$}&\textcolor{mygreen}{0.7~$\uparrow$}&\textcolor{mygreen}{1.0~$\uparrow$}&\textcolor{mygreen}{0.7~$\uparrow$}&\textcolor{mygreen}{0.7~$\uparrow$}&\textcolor{mygreen}{0.6~$\uparrow$} & \textcolor{mygreen}{0.6~$\uparrow$} \\
    
    \rowcolor{gray!20}LoRA$_{r=8}~\cite{TL:LoRA}$&0.44&73.7&94.1&72.4&99.4&91.3&85.6&56.3&86.8&95.9&85.2&73.4&83.6&64.6&51.4&79.5&85.7&51.5&35.1&46.5&76.5 & 74.3 \\

    \quad + ReSoRA&0.44&\textcolor{mygreen}{0.2~$\uparrow$}
    &\textcolor{red}{0.2~$\downarrow$}&\textcolor{mygreen}{0.5~$\uparrow$}&0.0&\textcolor{mygreen}{0.6~$\uparrow$}&\textcolor{mygreen}{1.9~$\uparrow$}&\textcolor{mygreen}{0.4~$\uparrow$}&\textcolor{mygreen}{0.9~$\uparrow$}&\textcolor{red}{0.2~$\downarrow$}&\textcolor{mygreen}{1.4~$\uparrow$}&\textcolor{mygreen}{1.8~$\uparrow$}&\textcolor{mygreen}{0.1~$\uparrow$}&\textcolor{red}{0.1~$\downarrow$}&\textcolor{mygreen}{0.6~$\uparrow$}&\textcolor{mygreen}{2.6~$\uparrow$}&\textcolor{mygreen}{0.8~$\uparrow$}&\textcolor{mygreen}{0.4~$\uparrow$}&\textcolor{mygreen}{0.6~$\uparrow$}&\textcolor{mygreen}{0.1~$\uparrow$}&\textcolor{mygreen}{0.7~$\uparrow$}&\textcolor{mygreen}{0.7~$\uparrow$}\\

    \rowcolor{gray!20}MoSLoRA$_{r=8}~\cite{TL:MSLoRA}$&0.40&73.5&94.1&71.6&99.2&91.1&85.6&56.2&87.5&95.5&85.9&76.3&82.7&64.5&52.4&81.2&86.1&53.9&33.2&46.7&76.8 & 74.6 \\

    \quad+ ReSoRA &0.40&\textcolor{mygreen}{0.6~$\uparrow$}&\textcolor{red}{0.6~$\downarrow$}&\textcolor{mygreen}{0.8~$\uparrow$}&\textcolor{mygreen}{0.1~$\uparrow$}&\textcolor{mygreen}{0.3~$\uparrow$}&\textcolor{mygreen}{0.9~$\uparrow$}&\textcolor{mygreen}{0.4~$\uparrow$}&\textcolor{mygreen}{0.4~$\uparrow$}&\textcolor{mygreen}{0.3~$\uparrow$}&\textcolor{mygreen}{0.1~$\uparrow$}&\textcolor{red}{0.3~$\downarrow$}&\textcolor{red}{0.1~$\downarrow$}&\textcolor{red}{0.3~$\downarrow$}&\textcolor{mygreen}{0.9~$\uparrow$}&\textcolor{mygreen}{0.8~$\uparrow$}&\textcolor{red}{0.1~$\downarrow$}&\textcolor{mygreen}{0.3~$\uparrow$}&\textcolor{mygreen}{1.1~$\uparrow$}&\textcolor{red}{0.5~$\downarrow$}&\textcolor{mygreen}{0.3~$\uparrow$}&\textcolor{mygreen}{0.2~$\uparrow$}\\
    
    \rowcolor{gray!20}DTL+~\cite{TL:DTL}&0.04&69.9&95.0&71.5&99.3&91.8&86.8&56.8&87.8&96.1&86.4&74.7&81.6&64.9&52.0&81.7&97.0&55.0&36.4&49.6&77.5 &75.5 \\

    \quad+ ReSoRA &0.04&\textcolor{mygreen}{0.7~$\uparrow$}&\textcolor{mygreen}{0.1~$\uparrow$}&0.0&\textcolor{mygreen}{0.1~$\uparrow$}&0.0&\textcolor{mygreen}{0.9~$\uparrow$}&\textcolor{red}{0.6~$\downarrow$}&\textcolor{mygreen}{0.5~$\uparrow$}&\textcolor{mygreen}{0.3~$\uparrow$}&\textcolor{mygreen}{0.8~$\uparrow$}&\textcolor{mygreen}{0.5~$\uparrow$}&\textcolor{red}{0.1~$\downarrow$}&\textcolor{mygreen}{0.4~$\uparrow$}&\textcolor{mygreen}{0.1~$\uparrow$}&\textcolor{mygreen}{1.0~$\uparrow$}&\textcolor{mygreen}{0.3~$\uparrow$}&0.0&\textcolor{mygreen}{0.3~$\uparrow$}&\textcolor{mygreen}{0.1~$\uparrow$}&\textcolor{mygreen}{0.4~$\uparrow$}&\textcolor{mygreen}{0.3~$\uparrow$}\\
	\specialrule{0em}{1pt}{1pt}
	\bottomrule[1.5pt]
	\end{tabular}
 	\label{tab:vtab-vit}
\end{table*}

\section{Experiments}
\label{sec:Experiments} 

% todo: 
\subsection{Experiments on Visual-Text Retrieval}
\noindent\textbf{Datasets.} 
We test ReSoRA on V\&L benchmarks, e.g., Flickr30K~\cite{Datasets:Flickr30k} on image-text retrieval, and MSR-VTT~\cite{Datasets:MSRVTT} on video-text retrieval.

\noindent\textbf{Implementation Details.} 
For image-text retrieval, we follow VSE$\infty$\cite{ITM:GPO}, employing a BERT-base encoder~\cite{TransF:BERT} for text and a Faster R-CNN~\cite{Detection:FasterR-CNN} as visual extractor.
For video-text retrieval, we adopt CLIP4Clip~\cite{VLP:CLIP4Clip} as the baseline with Text Transformer~\cite{TransF:GPT-2} and ViT-B/32~\cite{TransF:ViT} as the text and vision encoders.
We keep their default settings, e.g., choice of optimizer, warm-up schedule, input image resolution, video sequence length, input text processing, etc. 

\noindent\textbf{Main Results.}
Table~\ref{tab:VL_PETL} shows the comparisons on image-text and video-text retrieval. 
Notably, ReSoRA consistently outperforms LoRA methods, achieving superior retrieval performance across different downstream tasks. 
It further demonstrates the general effectiveness of ReSoRA in multi-modal architectures, where its plug-and-play design better captures and aligns visual-textual relationships. Consistent gains in image-text and video-text retrieval indicate that ReSoRA serves as an effective training regularizer for low-rank adaptation, without adding inference overhead.

\subsection{Experiments on VTAB-1K}
\noindent\textbf{Datasets.}
We further evaluate the effectiveness of ReSoRA on the VTAB-1K benchmark~\cite{Datasets:VTAB}, a comprehensive collection of 19 diverse visual classification datasets, each containing 1,000 training samples. These datasets are categorized into three distinct groups: 

(1) \textbf{Natural group}: This group includes natural image datasets from conventional cameras including CIFAR100~\cite{Datasets:Cifar100}, Caltech101~\cite{Datasets:Caltech101}, DTD~\cite{Datasets:DTD}, Flowers102~\cite{Datasets:Flowers102}, Pets~\cite{Datasets:Pets}, Sun397~\cite{Datasets:SUN}, and SVHN~\cite{Datasets:SVHN}. 

(2) \textbf{Specialized group}: This group covers imagery from specialized domains such as remote sensing and medical imaging, including Resisc45~\cite{Datasets:Resisc}, EuroSAT~\cite{Datasets:Eurosat}, Patch Camelyon~\cite{Datasets:PatchCamelyon}, and Diabetic Retinopathy~\cite{Datasets:Retinopathy}.

(3) \textbf{Structured group}: This group is designed to evaluate structured visual reasoning tasks, such as object counting and 3D orientation prediction. It includes CLEVR~\cite{Datasets:Clevr}, dSprites~\cite{Datasets:DSprites}, SmallNORB~\cite{Datasets:SmallNORB}, DMLab~\cite{Datasets:Dmlab}, and KITTI~\cite{Datasets:Kitti}.

\noindent\textbf{Implementation Details.} 
For VTAB-1K and few-shot experiments, we use ViT-B backbone~\cite{TransF:ViT} with LoRA variants under identical training settings: AdamW optimizer~\cite{Training:AdamW}, cosine learning rate decay, batch size 32, and weight decay $5\times10^{-2}$. We reproduce baselines with LoRA~\cite{TL:LoRA}, MoSLoRA~\cite{TL:MSLoRA}, FacT~\cite{TL:FacT}, and DTL~\cite{TL:DTL} for fair comparison. 
For LoRA and MoSLoRA, we set the rank $r=8$, while DTL and FacT use their original configurations. ReSoRA is applied with linear set-to-set regularization by default.

\begin{figure*}[htp]
    \centering
    \includegraphics[width=\linewidth]{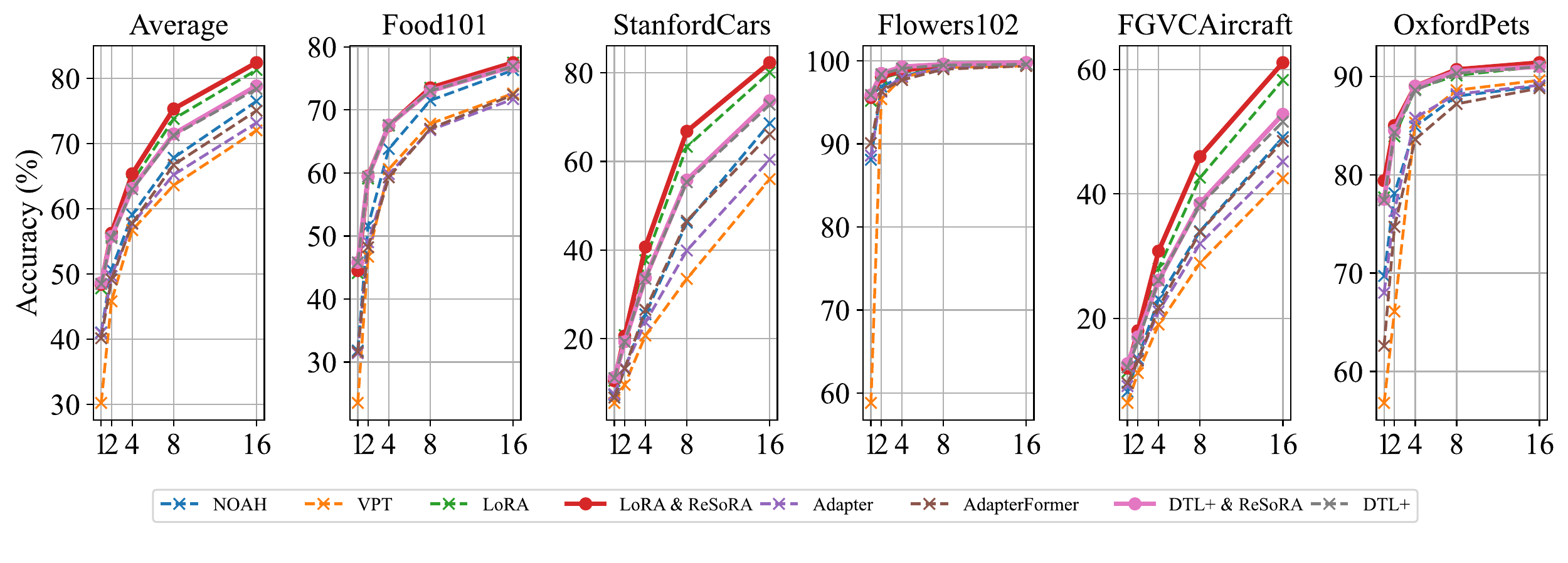}
    \caption{Few-shot results on five fine-grained recognition datasets. Solid lines indicate PETL methods with added ReSoRA; dashed lines show original results. All results are averaged over 3 seeds.}
    \label{fig:fewshot}
\end{figure*}

\noindent\textbf{Training Strategy.} 
As PETL methods are sensitive to parameter initialization, we adopt a two-stage training strategy to ensure stable convergence and reduce variance. Following prior work~\cite{TL:SPT}, models are trained for 100 epochs per dataset. In stage one, baseline PETL methods (e.g., LoRA, DTL, FacT) are trained without ReSoRA to obtain reference models; in stage two, ReSoRA with linear set-to-set regularization is introduced and trained for another 100 epochs, which is used across all experiments unless otherwise noted.

\noindent\textbf{Main Results.} 
We report top-1 accuracy for both the original models and our ReSoRA applied to various PEFT approaches, including various popular tuning methods such as LoRA~\cite{TL:LoRA}, MoSLoRA~\cite{TL:MSLoRA}, and FacT~\cite{TL:FacT}, as well as memory-efficient tuning methods like DTL~\cite{TL:DTL}, to validate ReSoRA’s effectiveness. We also compare against full fine-tuning (Full), partial tuning with a task-specific head (Linear), and competitive strategies such as BitFit~\cite{TL:BitFit}, VPT~\cite{TL:VPT}, LST~\cite{TL:LST}, AdaptFormer~\cite{TL:AdaptFormer}, NOAH~\cite{TL:NOAH}, and SSF~\cite{TL:SSF}. Finally, we include AdaLoRA~\cite{TL:AdaLoRA}, which reduces redundancy in LoRA by dynamically adjusting low-rank matrices during training.

\begin{table}[th]
\caption{Performance improvements on VTAB-1K using Swin Transformer and ConvNeXt pre-trained on ImageNet dataset.}
\begin{threeparttable}
	\centering
	\small
	\setlength{\tabcolsep}{1pt}
	\begin{tabular}{p{2.3cm}<{}p{1.5cm}<{\centering}p{1cm}<{\centering}p{1cm}<{\centering}p{1cm}<{\centering}p{1cm}<{\centering}}
	\toprule[1.5pt]
	Method &param(M)&Nat.&Spe.&Str.&Avg.\\\hline
	\specialrule{0em}{1pt}{1pt}
        % \multicolumn{5}{l}{Swin Transformer\tnote{*}~\cite{TransF:SwinTransformer}}
        \multicolumn{5}{l}{Swin Transformer~\cite{TransF:SwinTransformer}}
        \\\hline
        Full&86.7 &79.2&86.2&59.7&75.0\\
	Linear&0&73.5&80.8&33.5&62.6\\
	BitFit~\cite{TL:BitFit}&0.20&74.2&80.1&42.4&65.6\\
	VPT~\cite{TL:VPT}& 0.16&76.8&84.5&53.4&71.6\\
        \rowcolor{gray!20}
	FacT-TK~\cite{TL:FacT}& 0.14 & 83.4 & 86.8 & 61.3 & 77.2 \\
        \quad + ReSoRA  & 0.14 &\textcolor{red}{0.2~$\downarrow$}&\textcolor{mygreen}{0.4~$\uparrow$}&\textcolor{mygreen}{0.5~$\uparrow$}& \textcolor{mygreen}{0.2~$\uparrow$} \\ 
        \rowcolor{gray!20}
 	DTL+~\cite{TL:DTL} &0.09&82.2&86.5&65.7 & 78.1 \\ 
  
        \quad + ReSoRA &0.09&\textcolor{mygreen}{0.2~$\uparrow$}&\textcolor{mygreen}{0.6~$\uparrow$}&\textcolor{mygreen}{0.4~$\uparrow$} & \textcolor{mygreen}{0.4~$\uparrow$} \\ 
        \hline
       \multicolumn{5}{l}{ConvNeXt~\cite{CNN:convnet}}\\ \hline
       \rowcolor{gray!20} LoRA~\cite{TL:LoRA}&0.58&82.7&85.6&64.7&77.7\\
       
       \quad + ReSoRA  &0.58&\textcolor{mygreen}{0.4~$\uparrow$}&\textcolor{mygreen}{0.3~$\uparrow$}&\textcolor{mygreen}{0.2~$\uparrow$}& \textcolor{mygreen}{0.7~$\uparrow$} \\
	\bottomrule[1.5pt]
	\end{tabular}
        % \begin{tablenotes}
        % \item[*] We empirically discover that applying LoRA to Swin Transformers does not converge on the VTAB-1K benchmark.
        % \end{tablenotes}
\end{threeparttable}
\label{table:vtab-backbone}
\end{table}

The final results are summarized in Table~\ref{tab:vtab-vit}, presenting a comprehensive comparison between our proposed method and existing approaches on the VTAB-1K benchmark using the ViT-B/16 backbone.
The upper section of the table reports the performance of full fine-tuning and linear probing with a task-specific classification head. As shown, incorporating ReSoRA consistently enhances the baseline performance across multiple datasets without introducing additional parameters. Specifically, for LoRA, the addition of similarity regularization yields an average improvement of \textbf{0.7\%} across all datasets in the grouped categories, and a similar \textbf{0.7\%} gain in overall accuracy. 
We observe similar consistent improvements when applying ReSoRA to other PETL methods, including FacT, MoSLoRA, and DTL. These results further demonstrate its broad applicability across diverse fine-tuning strategies.

We further adopt Swin Transformer~\cite{TransF:SwinTransformer} and ConvNeXt~\cite{CNN:convnet} in Table~\ref{table:vtab-backbone} for conduct validation. Integrating ReSoRA consistently boosts the performance of these PETL methods across both Transformer and CNN architectures, demonstrating its versatility and robustness as a general enhancement for transfer learning.

\subsection{Experiments on Few-shot Learning}

\noindent\textbf{Datasets.}
We conduct few-shot learning on Food101~\cite{bossard2014food}, Stanford Cars~\cite{krause20133d}, Flowers102~\cite{1640927}, FGVCAircraft~\cite{maji2013fine}, and OxfordPets~\cite{Datasets:Pets}. These datasets are widely-used benchmarks for low-resource conditions and challenging fine-grained classification tasks.

\noindent\textbf{Implementation Details.}
We follow the settings in the VTAB-1K experiments including backbone, hyperparameters, GPU hardware, and two-stage training strategy, except that we vary the number of training samples from 1-shot to 16-shot. Results are averaged over three runs with different random seeds for reliability.

\begin{table}[t]
\caption{Performance improvements between feature-based and weight-based ReSoRA on VTAB-1k using ViT-B/16.}
	\centering
	\small
	\setlength{\tabcolsep}{1pt}
	\begin{tabular}{p{2.3cm}<{}p{1.5cm}<{\centering}p{1cm}<{\centering}p{1cm}<{\centering}p{1cm}<{\centering}p{1cm}<{\centering}}
	\toprule[1.5pt]
	Method &param(M)&Nat.&Spe.&Str.&Avg.\\\hline
	\specialrule{0em}{1pt}{1pt}
	Feature-Based&0.44&82.3&86.3&62.9 &77.2\\
	Weight-Based&0.44&82.2&86.1&62.5&76.9 \\
	\bottomrule[1.5pt]
	\end{tabular}\label{table:feature-weight}
\end{table}

\noindent\textbf{Main Results.}
Figure~\ref{fig:fewshot} presents a comparison of several PETL baselines, including LoRA, DTL+, Adapter, AdaptFormer, VPT, and NOAH, along with LoRA and DTL+ further enhanced by ReSoRA. 
The evaluation is conducted across five different datasets and shot settings (1, 2, 4, 8, 16), which test the methods under various levels of available training data. 
On average, ReSoRA yields a 0.5\%–1\% performance gain over baselines and continues to improve as training examples increase. These results demonstrate ReSoRA's effectiveness and robustness in few-shot settings and its ability to sustain gains with more data, highlighting its value in enhancing generalization and stability for parameter-efficient transfer learning.

\begin{table*}[htp]
\caption{Performance on VTAB-1K using ViT-B/16 with LoRA under different subspace regularizations.
Blue rows denote pairwise distance (Euclidean, Cosine), and green rows indicate set-to-set distance (Linear, Nonlinear).
Results are reported over 19 datasets grouped into Natural, Specialized, and Structured categories. Bold values indicate the best in each rank group.}
\centering
\setlength{\tabcolsep}{0.3pt}
\small
\begin{tabular}{p{2.2cm}<{}p{0.65cm}<{\centering}|p{0.65cm}<{\centering}p{0.65cm}<{\centering}p{0.65cm}<{\centering}p{0.65cm}<{\centering}p{0.65cm}<{\centering}p{0.65cm}<{\centering}p{0.65cm}<{\centering}|p{0.65cm}<{\centering}p{0.65cm}<{\centering}p{0.65cm}<{\centering}p{0.65cm}<{\centering}|p{0.65cm}<{\centering}p{0.65cm}<{\centering}p{0.65cm}<{\centering}p{0.65cm}<{\centering}p{0.65cm}<{\centering}p{0.65cm}<{\centering}p{0.65cm}<{\centering}p{0.65cm}<{\centering}|p{0.65cm}<{\centering}p{0.65cm}<{\centering}}
	\toprule[1.5pt]
	\multicolumn{2}{c|}{}&\multicolumn{7}{c|}{\textbf{Natural}}&\multicolumn{4}{c|}{\textbf{Specialized}}&\multicolumn{8}{c|}{\textbf{Structured}}&\\
	&\multicolumn{1}{c|}{\STAB{\rotatebox[origin=c]{90}{param (M)}}}
	&\multicolumn{1}{c}{\STAB{\rotatebox[origin=c]{90}{Cifar100}}}
	&\multicolumn{1}{c}{\STAB{\rotatebox[origin=c]{90}{Caltech101}}}
	&\multicolumn{1}{c}{\STAB{\rotatebox[origin=c]{90}{DTD}}}
	&\multicolumn{1}{c}{\STAB{\rotatebox[origin=c]{90}{Flower102}}}
	&\multicolumn{1}{c}{\STAB{\rotatebox[origin=c]{90}{Pets}}}
	&\multicolumn{1}{c}{\STAB{\rotatebox[origin=c]{90}{SVHN}}}
	&\multicolumn{1}{c|}{\STAB{\rotatebox[origin=c]{90}{Sun397}}}
	&\multicolumn{1}{c}{\STAB{\rotatebox[origin=c]{90}{Camelyon}}}
	&\multicolumn{1}{c}{\STAB{\rotatebox[origin=c]{90}{EuroSAT}}}
	&\multicolumn{1}{c}{\STAB{\rotatebox[origin=c]{90}{Resisc45}}}
	&\multicolumn{1}{c|}{\STAB{\rotatebox[origin=c]{90}{Retinopathy}}}
	&\multicolumn{1}{c}{\STAB{\rotatebox[origin=c]{90}{Clevr-Count}}}
	&\multicolumn{1}{c}{\STAB{\rotatebox[origin=c]{90}{Clevr-Dist}}}
	&\multicolumn{1}{c}{\STAB{\rotatebox[origin=c]{90}{DMLab}}}
	&\multicolumn{1}{c}{\STAB{\rotatebox[origin=c]{90}{KITTI-Dist}}}
	&\multicolumn{1}{c}{\STAB{\rotatebox[origin=c]{90}{dSpr-Loc}}}
	&\multicolumn{1}{c}{\STAB{\rotatebox[origin=c]{90}{dSpr-Ori}}}
	&\multicolumn{1}{c}{\STAB{\rotatebox[origin=c]{90}{sNORB-Azim}}}
	&\multicolumn{1}{c|}{\STAB{\rotatebox[origin=c]{90}{sNORB-Ele}}}
        &\multicolumn{1}{c}{\STAB{\rotatebox[origin=c]{90}{Average}}}
	&\multicolumn{1}{c}{\STAB{\rotatebox[origin=c]{90}{All Set Average}}}\\
	\specialrule{0em}{1pt}{1pt}
	\hline
	\specialrule{0em}{1pt}{1pt}
    LoRA$_{r=2}~\cite{TL:LoRA}$&0.11&74.9&93.2&73.3&99.3&91.4&84.9&57.4&86.8&95.2&84.7&74.9&82.8&64.2&52.2&80.7&84.4&51.7&32.1&46.0& 76.4 & 74.2 \\

    \rowcolor{blue!10}
    \quad + $R_e(B, A, X)$ &0.11&75.7&93.6&73.2&99.4&91.5&87.8 &57.5&87.7&95.3&85.6&74.2&83.4&64.7&53.3&80.2&84.9&52.4&32.7&46.4& \textbf{76.8} & \textbf{74.6} \\
    \rowcolor{blue!10}
    \quad + $R_c(B, A, X)$ &0.11&75.7&93.3&73.5&99.4&91.4&86.1&57.7&87.3&95.3&85.4&74.7&83.0&64.6&52.9&80.3&84.7&52.1&32.6&46.1& 76.7 & 74.5 \\
    \rowcolor{green!15}
    \quad + $R_n(B, A, X)$ &0.11&73.4&93.4&72.7&99.4&91.4&86.3 &57.7&87.0&95.2&85.7&73.2&83.1&64.6&53.3&80.7&84.9&52.1&32.3&45.5& 76.6 & 74.5 \\
    \rowcolor{green!15}
    \quad + $R_l(B, A, X)$ &0.11&75.7&93.4&73.5&99.5&91.4&86.2 &57.6&87.2&95.1&85.6&73.3&83.2&64.7&52.8&81.3&85.1&52.6&32.9&46.5& 76.7 & \textbf{74.6} \\

    % + CCA&0.11&-&-&-&-&-&- &-&-&-&-&-&-&-&-&-&-&-&-&-&-& -\\ 
    \hline
    LoRA$_{r=4}~\cite{TL:LoRA}$&0.22&74.5&93.9&71.4&99.3&91.4&83.6&56.8&87.2&95.5&85.7&75.3&83.2&65.0&52.3&81.3&85.8&52.4&33.0&46.0& 76.6 & 74.4 \\
        \rowcolor{blue!10}
    \quad + $R_e(B, A, X)$ &0.22&74.9&93.7&72.3&99.4&91.5&85.3 &56.9&88.0&95.9&86.4&75.4&83.3&65.5&52.9&81.6&86.8&53.0&34.0&45.6& \textbf{77.1} & \textbf{74.9} \\
    \rowcolor{blue!10}
    \quad + $R_c(B, A, X)$ &0.22&74.9&93.7&72.4&99.4&91.1&85.0&57.0&87.6&95.9&86.1&75.3&82.8&65.6&52.9&82.0&86.5&52.8&33.8&45.4 & 77.0 & 74.7 \\
    \rowcolor{green!15}
    \quad + $R_n(B, A, X)$ &0.22&75.2&93.6&72.1&99.4&91.6&85.3 &57.6&87.4&95.9&86.7&75.3&83.0&65.6&53.3&81.4&86.7&52.6&33.4&45.7& \textbf{77.1} & 74.8 \\
    \rowcolor{green!15}
    \quad + $R_l(B, A, X)$ &0.22&75.4&93.6&72.1&99.4&91.6&85.1 &57.2&87.7&96.0&86.4&75.3&83.1&65.5&53.2&81.7&86.8&53.0&33.8&45.2& \textbf{77.1} & \textbf{74.9} \\

    % + CCA&0.22&-&-&-&-&-&- &-&-&-&-&-&-&-&-&-&-&-&-&-&-& -\\ 
    \hline
    LoRA$_{r=8}~\cite{TL:LoRA}$&0.44&73.7&94.1&72.4&99.4&91.3&85.6&56.3&86.8&95.9&85.2&73.4&83.6&64.6&51.4&79.5&85.7&51.5&35.1&46.5& 76.5 & 74.3 \\
        \rowcolor{blue!10}
    \quad + $R_e(B, A, X)$ &0.44&73.6&93.1&72.2&99.2&91.0&87.4 &56.5&87.7&95.6&86.1&74.7&82.7&64.3&51.6&81.6&86.0&52.0&35.7&44.8& 76.7 & 74.5 \\
    \rowcolor{blue!10}
    \quad + $R_c(B, A, X)$ &0.44&73.9&93.3&72.7&99.3&91.4&87.0&56.7&87.5&95.7&86.3&73.5&82.7&64.7&51.4&80.6&85.7&51.7&35.6&46.1& 76.7 & 74.5 \\
    \rowcolor{green!15}
    \quad + $R_n(B, A, X)$&0.44& 74.0 &93.9&72.7&99.3&91.7&86.7 &56.7&87.8&95.8&86.5&74.2&82.9&64.5&51.4&80.9&85.8&52.7&35.3&46.1& 76.8 & 74.6 \\
    \rowcolor{green!15}
    \quad + $R_l(B, A, X)$& 0.44 & 73.9 & 93.9 &72.9&99.4&91.9&87.5 &56.7&87.7&95.7&86.6&75.2&83.7&64.5&52.0&82.1&86.5&51.9&35.7&46.6& \textbf{77.2} & \textbf{75.0} \\

	\specialrule{0em}{1pt}{1pt}
	\bottomrule[1.5pt]
	\end{tabular}
 	\label{tab:vtab-vit-different-regularization}
\end{table*}

\begin{figure*}[htp]
    \centering
    \includegraphics[width=0.95\linewidth]{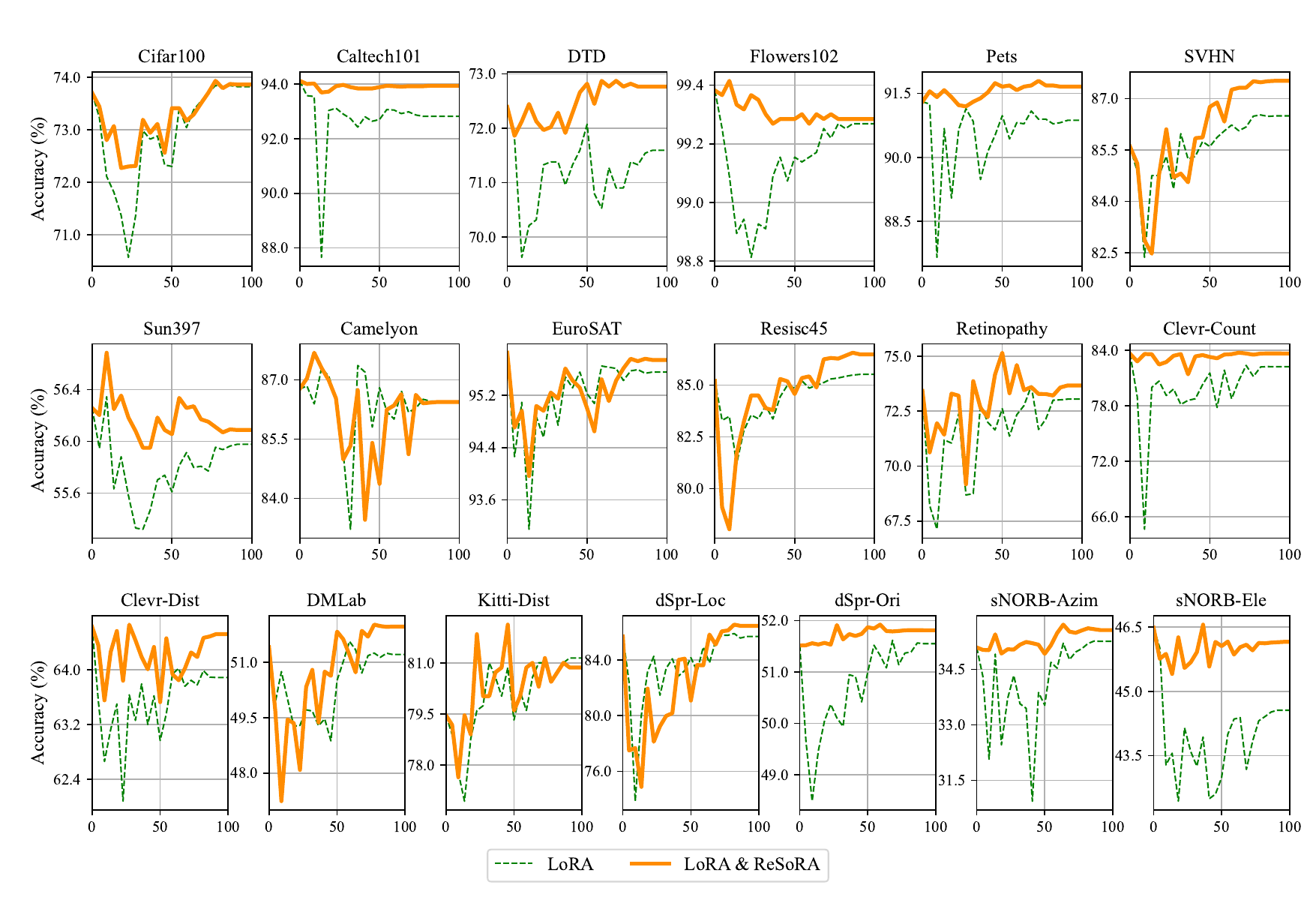}
    \caption{Effective training by adding ReSoRA in the second stage. The curves show test accuracy across datasets during training.}
    \label{fig:accuracy_comparison}
\end{figure*}

\section{Ablation Studies}

\subsection{Weight vs. Feature Regularization}
To compare the effectiveness of applying the same regularization at different levels, we evaluate linear set-to-set similarity regularization when applied to either the weight subspace of low-rank adaptation or the resulting feature representations.

As shown in Table~\ref{table:feature-weight}, although applying regularization directly to the LoRA weight matrices helps mitigate redundancy within the parameter space, it consistently underperforms when compared to regularization applied at the feature level. This performance gap suggests a fundamental limitation of weight-level constraints: they influence the learned representations only implicitly and may fail to align with the underlying task semantics. In contrast, feature-level regularization provides a more direct and effective mechanism for shaping the output representations, enabling the model to better capture task-relevant variations. As a result, it leads to improved generalization and adaptability across downstream tasks.

\subsection{Comparisons of Different Regularization}
\label{Different Similarity Regularization}
To better understand the role of different regularizations, we conduct a comprehensive study on the ViT-B/16 backbone with LoRA of varying ranks ($n = 2, 4, 8$). We compare pairwise distance measures (e.g., Euclidean and cosine distance) with set-to-set approaches based on Linear and Nonlinear distance formulations.

As shown in Table~\ref{tab:vtab-vit-different-regularization}, applying pairwise regularization techniques, such as Euclidean distance and cosine distance, to the incremental features proves to be effective in reducing redundancy across various LoRA ranks. Notably, even in the low-rank setting ($r=2$), where the rank is minimal, significant redundancy persists. This finding suggests that the overlap of subspaces is not solely a consequence of high-rank configurations. Instead, it highlights that subspace redundancy is an inherent issue that can occur across all rank configurations, not just in higher-dimensional settings. This indicates that the redundancy problem is fundamental and needs to be addressed regardless of the rank dimension.

At higher ranks (e.g., $n=4$, $n=8$), the performance gains of pairwise regularization become more pronounced, particularly on datasets such as SVHN and KITTI. Besides, the set-to-set regularization, including both linear and nonlinear items, further improves performance by encouraging more compact and discriminative feature subspaces. Compared to pairwise methods, set-to-set regularization shows greater robustness to input variations and consistently yields stronger results in more complex scenarios.

Overall, both regularization forms play a crucial role in reducing redundancy within LoRA parameters and boosting adaptation performance. Among all the strategies evaluated, the linear set-to-set item stands out by delivering the most significant improvements, indicating its superior ability to enforce diversity and enhance the representational capacity of the learned subspaces.

\begin{table}[th]
\caption{Memory and resource analyses of different ReSoRA variants on VTAB-1K using ViT-B/16 backbone.}
\centering
\small
\setlength{\tabcolsep}{6pt}
\begin{tabular}{lcccc}
\toprule[1.5pt]
Method & Mem (GB) & FLOPs (T) & Time (\%) & Acc (\%) \\
\midrule
Full & 4.9 & 0.54 & - & 68.9 \\
\rowcolor{gray!20} LoRA~\cite{TL:LoRA} & 3.4 & 0.56 & 100 & 76.4 \\
\quad + $R_e(B, A, X)$ & 3.4 & 0.83 & +18 & 0.4~$\uparrow$ \\
\quad + $R_c(B, A, X)$ & 4.0 & 0.83 & +32 & 0.3~$\uparrow$ \\
\quad + $R_n(B, A, X)$ & 4.0 & 0.83 & +34 & 0.2~$\uparrow$ \\
\quad + $R_l(B, A, X)$ & 4.0 & 0.83 & +34 & 0.3~$\uparrow$ \\
\rowcolor{gray!20} DTL~\cite{TL:DTL} & 2.4 & 0.56 & 100 & 77.5 \\
\quad + $R_e(B, A, X)$ & 2.5 & 0.65 & +14 & 0.2~$\uparrow$ \\
\quad + $R_c(B, A, X)$ & 3.0 & 0.65 & +32 & 0.3~$\uparrow$ \\
\quad + $R_n(B, A, X)$ & 3.0 & 0.65 & +32 & 0.4~$\uparrow$ \\
\quad + $R_l(B, A, X)$ & 3.0 & 0.65 & +32 & 0.4~$\uparrow$ \\
\bottomrule[1.5pt]
\end{tabular}
\label{tab:efficiency-analysis}
\end{table}

\subsection{Memory and Resource Usage}

To thoroughly assess ReSoRA’s efficiency, we compare memory usage, computational cost (FLOPs/iteration), training time, and accuracy across different forms, evaluating pairwise (Euclidean: $R_e$, Cosine: $R_c$) and set-to-set (Linear: $R_l$, Nonlinear: $R_n$) variants.

Table~\ref{tab:efficiency-analysis} summarizes the results. Compared to baseline LoRA~\cite{TL:LoRA} and DTL~\cite{TL:DTL}, our ReSoRA consistently improves accuracy by approximately 0.2\% to 0.4\% across different configurations. In particular, these performance improvements are achieved with minimal additional memory usage (at most 0.6 GB) and a modest increase in training time (ranging from 14\% to 34\%), while introducing no extra inference overhead. Although the linear set-to-set form ($R_l$) generally achieves better performance, the Euclidean pairwise form ($R_e$) offers a balance between accuracy and computational efficiency.

\subsection{Effective Training of ReSoRA} 

To validate the training configurations for the proposed ReSoRA, we perform ablation studies using the ViT-B/16 backbone with trainable LoRA adapters on the VTAB-1k benchmark. As illustrated in Figure~\ref{fig:accuracy_comparison}, we compare the top-1 accuracy during the second stage of training between models optimized with the original objective and those enhanced by our regularization strategy. 

% As shown in Figure~\ref{fig:accuracy_comparison}, the models trained solely with the original loss exhibit a noticeable decline in accuracy during the second stage, indicating the overfitting problem. In contrast, introducing our proposed regularization in this stage not only mitigates this degradation but also improves performance beyond the baseline. On datasets such as Caltech101\cite{Datasets:Caltech101}, Pets~\cite{Datasets:Pets}, Clevr-Count~\cite{Datasets:Clevr}, dSpr-Ori~\cite{Datasets:DSprites}, and sNORB-Azim~\cite{Datasets:SmallNORB}, the regularized models achieve higher accuracy early in training and converge more rapidly. For more challenging datasets like Resisc45~\cite{Datasets:Resisc}, DMLab~\cite{Datasets:Dmlab}, and dSpr-Loc~\cite{Datasets:DSprites}, the performance improves steadily as the subspace is progressively optimized, ultimately exceeding the baseline. 
% Notably, our proposed ReSoRA serves as a post-stage training strategy, which is the default configurations in all our experiments.

In Figure~\ref{fig:accuracy_comparison}, models trained solely with the original loss function experience a noticeable drop in accuracy during the second training stage, which reflects an overfitting issue caused by insufficient regularization. In contrast, our ReSoRA not only mitigates this degradation but also pushes performance beyond the baseline levels. On datasets such as Caltech101~\cite{Datasets:Caltech101}, Pets~\cite{Datasets:Pets}, Clevr-Count~\cite{Datasets:Clevr}, dSpr-Ori~\cite{Datasets:DSprites}, and sNORB-Azim~\cite{Datasets:SmallNORB}, the regularized models exhibit faster convergence and achieve higher accuracy early in training, demonstrating the strong stabilizing effect of ReSoRA.

For more challenging datasets, including Resisc45~\cite{Datasets:Resisc}, DMLab~\cite{Datasets:Dmlab}, and dSpr-Loc~\cite{Datasets:DSprites}, we discover that performance gains continue to improve steadily as the subspace is progressively refined, ultimately surpassing the baseline results. These observations underscore the effectiveness of ReSoRA as a post-stage training strategy, which is adopted as the default configuration across all our experiments to ensure consistent gains and robust generalization.

\section{Conclusion}
\label{sec:Conclusion}
In this paper, we propose \textbf{ReSoRA}, a simple yet highly effective method for reducing parameter redundancy in low-rank adaptation. Specifically, ReSoRA works by decomposing the output space of the adaptive branch into multiple subspaces, allowing for more targeted and efficient adaptation. To encourage diversity among these subspaces, ReSoRA introduces a regularization term that penalizes similarity across them, fostering better feature representation and enhancing the model’s adaptability. Extensive experiments on vision-text retrieval and visual classification tasks demonstrate that our ReSoRA consistently improves various baseline models and application scenarios, particularly in low-data and few-shot settings. Moreover, it integrates seamlessly with existing LoRA-like methods without adding inference overhead. These results validate its effectiveness and establish it as a promising, general solution for efficient fine-tuning across diverse domains and tasks.

\noindent \textbf{Discussion.}
We present ReSoRA as a novel, effective, and easy-to-implement regularization technique for low-rank adaptation across diverse backbones and application domains. A current limitation is its role as a post-training refinement step, which adds extra training time. In the future, we plan to integrate ReSoRA into larger-scale foundation models and generative AI pipelines as better computational resources and high-quality data become available.

\noindent \textbf{Acknowledgements.}
This paper is supported in part by the National Natural Science Foundation of China under Grant Nos.~62441231 and 62293542, the Liaoning Province Science and Technology Plan under Grant No.~2023JH26/10200016, the Dalian City Science and Technology Innovation Fund under Grant No.~2023JJ11CG001, the Natural Science Foundation of Zhejiang Province under Grant No.~LD25F020001, and the Ningbo Key R\&D Project under Grant No.~2025Z039, and in part by the Grant 2024-0011(ZX20240867).

\bibliographystyle{ACM-Reference-Format}
\balance
\bibliography{sample-base}

% \appendix
% \clearpage 
% \input{appendix.tex}

\end{document}